# A Distance-Based Branch and Bound Feature Selection Algorithm

**Ari Frank***
Computer Science Dept.,
Technion, Haifa 32000, Israel.
*arf@cs.ucsd.edu*

**Dan Geiger**
Computer Science Dept.,
Technion, Haifa 32000, Israel.
*dang@cs.technion.ac.il*

**Zohar Yakhini**
Computer Science Dept.,
Technion, Haifa 32000, Israel
and Agilent Laboratories.
*zohar_yakhini@agilent.com*

## Abstract

There is no known efficient method for selecting $k$ Gaussian features from $n$ which achieve the lowest Bayesian classification error. We show an example of how greedy algorithms faced with this task are led to give results that are not optimal. This motivates us to propose a more robust approach. We present a *Branch and Bound* algorithm for finding a subset of $k$ independent Gaussian features which minimizes the naive Bayesian classification error. Our algorithm uses additive monotonic distance measures to produce bounds for the Bayesian classification error in order to exclude many feature subsets from evaluation, while still returning an optimal solution. We test our method on synthetic data as well as data obtained from gene expression profiling.

## 1 Introduction

Feature selection is an essential step to enhance correct classification in the presence of many irrelevant features, or when the statistical model cannot be estimated accurately due to a small number of training samples. When the probability distributions $p(\vec{x}|\omega_i)$ of a vector $\vec{x}$ of measured features given class $\omega_i$ are known exactly, removing some measured features of $\vec{x}$ cannot reduce the classification error (Van Campenhout, 1982). However, in practice, $p(\vec{x}|\omega)$ is never known precisely. Consequently, choosing a subset of features of $\vec{x}$ often improves the error rate. This is apparent in many test cases examined in the literature, where the removal of redundant, irrelevant or correlated features improved the performance of various classifiers (Langley & Sage, 1994; Kohavi & John, 1997; Koller & Sahami, 1996). In addition, the computational effort required for classification usually grows with the number of features, thus feature selection can reduce the running time (Koller & Sahami, 1996). A review of many of the popular feature selection methods is given by Dash & Liu (1997).

The question at hand is how to choose a subset of $k$ features out of $n$ that guarantees the lowest classification error. No algorithm is known that finds the optimal feature subset without, in the worst case, exhausting all $\binom{n}{k}$ possible subsets; an insurmountable task for common values of $k$ and $n$. A result due to Cover & Van Campenhout (1977) even states that in a specific domain they describe, any non exhaustive feature selection algorithm, which selects subsets according to their Bayesian classification error, can be made to perform arbitrarily bad.

We present herein an approach that is capable of drastically reducing the number of different feature subsets that need to be evaluated while searching for an optimal subset. Using bounds on the Bayesian classification error, we prune subsets of features that are no longer candidates for having the lowest error, given the subsets examined so far. Our algorithm uses a *Branch and Bound* technique, in which the state space tree for the problem at hand is explored. At each point of searching the tree, a bound is computed for the best solution possible in the current subtree. Promising nodes in the tree are expanded, whereas nodes for which the lower bound is larger than the best solution found so far, are pruned. We compare our algorithm to another Branch and Bound feature selection algorithm due to Narendra & Fukunaga (1977) in Section 6.2.

This paper is organized as follows: in Section 2 we define the statistical model, followed by Section 3 in which we describe the Bhattacharyya distance bound. An example of the shortcomings of greedy feature selection is given in Section 4. In Section 5 we present the subset tree data structure used in our algorithm. We describe the algorithm itself in Section 6. In Section 7 we analyze its performance, including an application to gene expression data. We end with a short conclusion in Section 8.

## 2 The Statistical Model

Let $f_{\mu,\sigma}$ denote the pdf of a Normal distribution with a mean $\mu$ and a standard deviation $\sigma$: $f_{\mu,\sigma}(x) = 1/(\sqrt{2\pi}\sigma)\exp(-\frac{(x-\mu)^2}{2\sigma^2})$. The classification problem we address is defined via two independent $n$ dimensional multivariate normal distributions, $p(\vec{x}|\omega_i) = \prod_{j=1}^{n} f_{\mu_{ij}, \sigma_{ij}}(x_j)$, $i = 1, 2$. The classification is performed by a *naive Bayesian classifier*. Given a point $\vec{x} = (x_1, \ldots, x_n)$, called a *feature vector*, conditional probabilities $p(\vec{x}|\omega_i)$, called the likelihood functions, and class priors $p_{\omega_1}, p_{\omega_2}$, the classifier decides to which class a point $\vec{x}$ is more likely to belong. It does so by assigning $\vec{x}$ to the class which maximizes the posterior probability, given by Bayes' rule:

$$p(\omega_i|\vec{x}) = \frac{p(\vec{x}|\omega_i) \cdot p_{\omega_i}}{p(\vec{x})} = \frac{p(\vec{x}|\omega_i) \cdot p_{\omega_i}}{p(\vec{x}|\omega_1) \cdot p_{\omega_1} + p(\vec{x}|\omega_2) \cdot p_{\omega_2}}.$$

---

* Current address: Dept. of Computer Science and Engineering, University of California San Diego, 9500 Gilman Dr., La Jolla, Ca., 92093-0114.

The prediction error using this decision rule when the parameters $\mu_{ij}, \sigma_{ij}$ are known is referred to as the *Bayesian classification error*:

$$P_e = \int_{\Re^n} \min[p_{\omega_1} \cdot p(\vec{x}|\omega_1), \ p_{\omega_2} \cdot p(\vec{x}|\omega_2)] \, d\vec{x} \quad (1)$$

When the parameters $\mu_{ij}, \sigma_{ij}$ are known, the Bayesian classifier has been proven to be optimal relative to the Bayesian classification error (Fukunaga, 1990). When the parameters are unknown, as in most practical applications, the terms $p(\vec{x}|\omega_i)$ are replaced with their estimate $\hat{p}(\vec{x}|\omega_i)$ computed from the training data. The naive Bayesian classifier, despite its simplicity, performs well across a large range of datasets, often outperforming sophisticated classifiers (Domingos & Pazzani, 1997).

## 3 Distance Based Error Bounds

There are several types of upper and lower bounds on the Bayesian error of classification (Ben-Bassat, 1982). Some of these bounds are based on distance measures between the distributions $p(x_S|\omega_1)$ and $p(x_S|\omega_2)$, where the vector $x_S$ is the part of $\vec{x}$ with indices in $S \subseteq \{1, \ldots, n\}$. We define the distance of two classes $\omega_1$ and $\omega_2$ with respect to a subset of features $x_S$ to be the distance between $p(x_S|\omega_1)$ and $p(x_S|\omega_2)$, denoted by $Dist(S)$. A high distance is indicative of well separated distributions. We focus on distance measures having three properties:

1. **Closed Form** - A distance that is simple to calculate. The bound based on the distance measure must be easier to calculate than the Bayesian error $P_e$.

2. **Additive** - The distance of a subset of independent features equals the sum of distances of the individual features.

3. **Monotonic** - For every two subsets of features $x_S, x_T$, if $T \subseteq S$ then $Dist(T) \leq Dist(S)$.

We focus on the *Bhattacharyya* distance (Fukunaga, 1990), a special case of the *Chernoff* distance, which under the statistical model of independent multivariate normal distributions, satisfies the above conditions. We do not use the more common divergence (KL) distance because its bounds are less tight than the Bhattacharyya distance, affecting the algorithm's efficiency. We use the terms $p(x_S|\omega_i)$ rather than their estimate $\hat{p}(x_S|\omega_i)$ in the exposition of the bounds and of our algorithm. We discuss the estimation of $p(x_S|\omega_1)$ in Section 6.1.

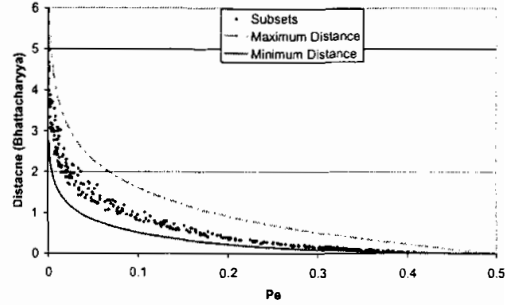

Figure 1: Plot of $P_e$ and the Bhattacharyya distances of 400 subsets of 5 features with parameters uniformly drawn [0,1]. The theoretical Bhattacharyya upper and lower distance bounds of Eq.(4) are also shown.

The error bounds described below, in particular the lower distance bound on the Bayesian classification error which is required by our algorithm, are defined for the Bhattacharyya distances of distributions containing two classes. There is no formal definition of these distances in domains with more than two classes. Ben-Bassat (1982) suggests the distance of multi-class features be the weighted sum of the distances of the pairs of classes. However the use of such a distance in the bounds described below is not valid, therefore when more than two classes are involved the algorithm becomes a heuristic.

The Bhattacharyya bound is based on the fact that the integrand in the Bayesian error from Eq.(1) can be bounded with a simple term. When the two joint conditional distributions of features are multivariate normal, say $N(\mu_1, \Sigma_1)$ and $N(\mu_2, \Sigma_2)$, their *Bhattacharyya distance* $B$ can be defined by (Fukunaga, 1990)

$$\frac{1}{8}(\mu_2 - \mu_1)' \left[\frac{\Sigma_1 + \Sigma_2}{2}\right]^{-1} (\mu_2 - \mu_1) + \frac{1}{2} \ln \frac{|\frac{\Sigma_1 + \Sigma_2}{2}|}{\sqrt{|\Sigma_1||\Sigma_2|}} \quad (2)$$

For independent features, where the covariance matrices $\Sigma_i$ are diagonal, Eq.(2) simplifies to

$$B = \sum_{i=1}^{n} \left\{ \frac{1}{4} \frac{(\mu_{2i} - \mu_{1i})^2}{\sigma_{1i}^2 + \sigma_{2i}^2} + \frac{1}{2} \ln \frac{\sigma_{1i}^2 + \sigma_{2i}^2}{2\sigma_{1i} \cdot \sigma_{2i}} \right\} \quad (3)$$

Eq.(3) has an easy to calculate closed-form. The Bhattacharyya distance is always positive. It is additive by definition, and therefore monotonic. Using the Bhattacharyya distance, the following bounds on the Bayesian error $P_e$ have been derived (Devijver & Kittler, 1982, p. 58):

$$\frac{1}{2}(1 - \sqrt{1 - 4p_{\omega_1} p_{\omega_2} \, e^{-2B}}) \leq P_e \leq \sqrt{p_{\omega_1} p_{\omega_2}} \, e^{-B} \quad (4)$$

We denote by THRESH($P_e$) the distance $B$ obtained when $P_e$ equals the lower bound in Eq.(4):

$$P_e = \frac{1}{2}(1 - \sqrt{1 - 4p_{\omega_1} p_{\omega_2} e^{-2B}}), \quad \Rightarrow$$

$$\text{THRESH}(P_e) = B = \frac{1}{2} \ln \frac{4 p_{\omega_1} p_{\omega_2}}{1 - (1 - 2p_e)^2} \quad (5)$$

The quantity THRESH($P_e$) is the lowest possible distance for which a subset of features $x_S$ can still have a Bayesian error lower than $P_e$. This is the threshold value we refer to in our feature selection algorithm. Figure 1 depicts the tightness of the lower and upper bounds of Eq.(4). Only the lower bound is used by our algorithm.

## 4 Greedy Selection is Not Optimal

We first give an example of an optimal feature subset that does not include any of the features with the individual lowest Bayesian classification errors. This means that greedy selection of features according to their classification error does not give optimal subsets. Therefore a different approach for feature selection is needed.

Table 1: Parameters Of Feature Types.

| Type | Class 1 | Class 2 | $P_e$ |
|---|---|---|---|
| A | $N(0, 1^2)$ | $N(-2.0254, 1.3946^2)$ | 0.1945 |
| B | $N(0, 1^2)$ | $N(0.9396, 0.4045^2)$ | 0.2076 |

Consider two classes with prior probabilities $p_{\omega_1} = p_{\omega_2} = 0.5$ and two types of features with the parameters as in Table 1. Consider 10 features, 5 features of type $A$ and 5 of type $B$. Which is the subset of 5 features that has the lowest classification error? If the features are chosen according to their individual classification errors, the subset with 5 features of type $A$ is selected since $P_e(A) < P_e(B)$. A subset with 5 type $A$ features has a $P_e$ of 0.0253. However, a subset with 5 features of type $B$ has a lower $P_e$ value of 0.0229. Figure 2 depicts Bayesian classification error rates for different possible subsets of one to five features.

A *Forward Sequential Search* algorithm (FSS) is another greedy feature selection approach. It starts with an empty set of features, and in each iteration it greedily adds the most promising feature. In our example, the FSS algorithm starts with the subset containing a single feature $A$, and in each iteration it adds another $A$, since adding an $A$ to a pure $A$ subset is always the most beneficial step. It proceeds until it reaches the subset $AAAAA$, which is not optimal.

An FSS algorithm with $r$ replacements starts with the best subset of $r$ features. At each iteration the algorithm adds the feature that yields the greatest reduction in the classification error. It then considers to replace up to $r$ of the features in that subset with any of the features not selected yet. Note that this operation adds to the running complexity of the algorithm an order of $O(n^{2r})$.

Figure 2 shows that even an FSS algorithm with 2 replacements returns a subset which is not optimal. Such an algorithm starts with the subset $AA$, since $P_e(AA)$ is the lowest error of all subsets of two features. It then adds an $A$ to form the subset $AAA$. It does not change this subset since any replacement of two features gives a higher classification error rate. For this reason it keeps on adding $A$'s until it selects the subset $AAAAA$, which is not optimal. This example shows how an FSS algorithm that performs a limited local search can return a suboptimal results. For simplicity we showed an example of how an algorithm with $r = 2$ replacements fails. Examples can be constructed where an FSS algorithm with $r > 2$ replacements also fails.

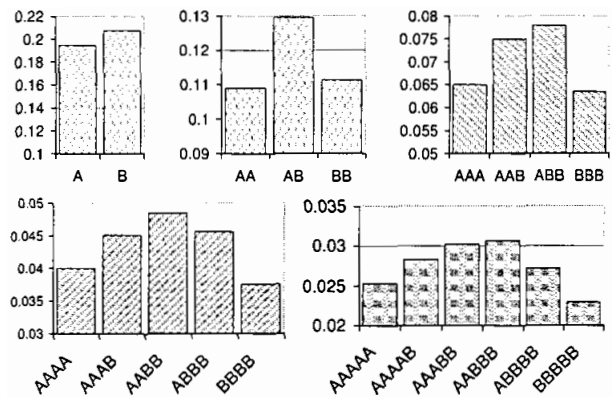

Figure 2: The classification errors of different subsets of features from both types.

A partial explanation for this phenomenon where using features that have individual lowest classification errors yields a suboptimal feature subset is given when the distances of the features are examined. Although type $A$ features have lower $P_e$, the *divergence* distance (Devijver & Kittler, 1982) of type $B$ features is greater than that of type $A$ (the divergence distance of type $A$ is 1.67, and the divergence distance of type $B$ is 2.64). Therefore, the larger the subset of features, the greater is the distance of a subset type $B$ features, compared to the distance of a subset of $A$'s. Generally, when the divergence distance yields a different ordering on the features than the ordering induced by $P_e$, a subset of $n$ features of the type with the higher $P_e$ will yield a lower error than a subset of $n$ features with the lower $P_e$, for a suitably large $n$ (Cover, 1974).

## 5 The Subset Tree Representation

One way to represent all possible feature subsets of size $k$, from a pool of $n$ features, is through a subset tree. A *subset tree* is an ordered directed tree where each node is denoted by a unique subset of $\{1, \ldots, n\}$ and each edge has a non-unique label from $1, \ldots, n$. A subset tree has $k$ levels of nodes besides the root $\emptyset$ which is on level 0. The edges leaving each non-terminal node are labeled as follows: if the node is on level $j$, $0 \leq j < k$, and the edge entering it has label $m$ (for the root $m = 0$), the edges leaving it are labeled $m+1, \ldots, n-k+j+1$ from left to right. The unique subset associated with a node $v$ is the set of the labels of edges on the path from the root to $v$. Figure 3 depicts a subset tree.

The feature selection algorithm we present uses a subset's distance to determine if it is a possible candidate for being the subset with the lowest Bayesian classification error. Subsets with a distance below some threshold, need not be examined, so the subset tree is pruned. The monotonicity and additivity of the distance enable efficient pruning.

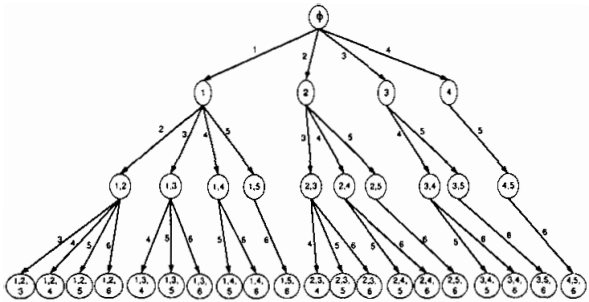

Figure 3: The subset tree for $n = 6$, with $k = 3$ levels. Every node is associated with a subset of $\{1,\ldots,n\}$. Level $k$ in the tree has $\binom{n}{k}$ nodes representing all possible subsets of size $k$ from $n$.

In order to represent subsets of features via a subset tree, features are first sorted according to decreasing distances, and indexed in that order (Feature 1 with the highest distance, Feature $n$ with the lowest). Every node $S$, at level $j$ in the subset tree, represents a feature subset $x_S$, which includes the features whose indices appear in $S = \{s_1, s_2, \ldots, s_j\}$. Let $Dist(S)$ denote the distance of the subset of features $x_S$. Since additive distances are used, $Dist(S)=\sum_{i=1}^{j} Dist(\{s_i\})$. Two nodes in the subset tree are *siblings* if they have the same parent node.

**Proposition 1:** *Let $S_1$ and $S_2$ be two direct children of $S$. If $S_1$ is to the left of $S_2$, then $Dist(S_1) \geq Dist(S_2)$.*

**Proof:** We have $Dist(S_i)=Dist(S)+Dist(S_i \setminus S)$, $i = 1, 2$. Since the index added to $S_1$ is lower than the index added to $S_2$, it follows that $Dist(S_1 \setminus S) \geq Dist(S_2 \setminus S)$.

**Implication for pruning:** When terminal nodes are examined, if a node has a distance below a threshold, the examination of its right siblings is not needed because these nodes must also have a distance below the threshold.

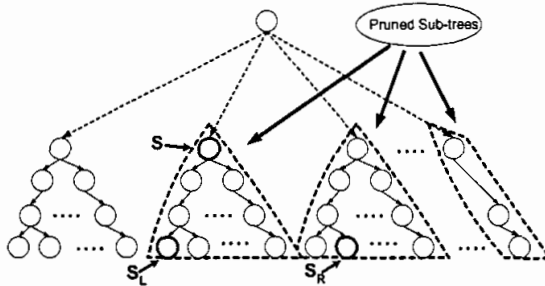

Figure 4: Pruning according to Proposition 2. If $S_L$ has a distance below the threshold, the marked subtrees are pruned.

For a non-terminal node $S$ in a subset tree, let $S_L$ denote the left-most terminal node in the subtree rooted at $S$. Let $S_R$ denote a terminal node which is to the right of $S_L$ in a subtree rooted at $S$, or one of $S$'s right siblings. See example in Figure 4.

**Proposition 2:** *For every non-terminal node $S$ in the subset tree, $Dist(S_L) \geq Dist(S_R)$.*

**Proof:** $S_L$ and $S_R$ share a common path from the root to some node $S'$. Therefore $Dist(S_i) = Dist(S') + Dist(S_i \setminus S')$, $i =$L,R. From $S'$ the path to $S_R$ branches to the right, while the path to $S_L$ stays left. After the branching, at each level, the indices on the path to $S_R$ are greater than the indices on the path to $S_L$. Therefore $Dist(S_L \setminus S') \geq Dist(S_R \setminus S')$, and the claim follows.

**Implication for pruning:** For every node $S$ in the subset tree, if $S$'s leftmost terminal node has a distance below the threshold, so will all the terminal nodes in $S$'s subtree, or in any of the subtrees rooted at $S$'s right siblings. These subtrees can be pruned, as illustrated in Figure 4.

## 6 The Algorithm

To find the subset of $k$ features with the lowest $P_e$ our algorithm searches the subset tree in a depth-first manner. While examining each terminal node's $P_e$, a minimum distance threshold THRESH($P_e$) is kept. This threshold, which is a function of the lowest $P_e$ encountered by the algorithm to that point, is the minimal distance needed of any terminal subset in order for it to be considered by the algorithm. Using this threshold along with propositions 1 and 2, the subset tree is pruned, reducing the number of subsets examined. The algorithm is described in Figure 5.

Note that the subset tree is not completely expanded by the algorithm. Only the path to the current subset being examined is saved. Therefore, the space complexity of the algorithm is $O(k)$.

Before running the algorithm, the $n$ features are sorted according to decreasing distances. The algorithm is initially called with *scan_tree*($\emptyset$). Throughout the scanning of the subset tree, the subset with the lowest $P_e$ is stored in $S_A$, and the lowest classification error is stored in $min\_P_e$. This classification error defines a minimum distance threshold, which is computed by the function THRESH according to Eq.(5), and is stored in $min\_dis$. $Dist(S)$ is the distance of the features whose indices are in $S$, as defined in Section 3. The Boolean value returned indicates whether the subtree rooted at $S$ is pruned.

The algorithm's execution depends on the size of $S$:

**Terminal Step: Evaluation of terminal nodes**
When $|S| = k$, the scanning of the subset tree does not go deeper. Depending on $Dist(S)$, two possible actions are taken:

If $Dist(S) < min\_dis$, $S$ is not evaluated. In addition, the value *true* is returned, to indicate that $S$ was pruned, and to stop the evaluation of $S$'s right siblings (Proposition 1).

If $Dist(S) \geq min\_dis$, $P_e(S)$ is computed. If it is lower than the current $min\_P_e$, the subset $S$ and its error are recorded. The new (higher) $min\_dis$ is computed accordingly. The value *false* is returned because $S$ was not pruned. Examination of its right siblings continues.

**Recursive Step: Evaluation of inner-nodes**
If $S$ is not a terminal node, namely $|S| < k$, *scan_tree* is called on each of $S$'s children. If $S$'s leftmost child is not pruned, *false* is returned to indicate that $S$'s next sibling should be checked. If $S$'s leftmost child is pruned, the value *true* is returned to indicate that $S$'s right siblings should not be checked (Proposition 2).

```
Feature Selection Algorithm
Initialization:
S_A ← ∅    // current best subset of k features
min_P_e ← 1 // classification error of S_A
min_dis ← 0 // minimum distance threshold
n ← Total number of features
k ← Desired subset size
Boolean scan_tree(S)
Initial call: scan_tree(∅).
Input: A subset of features S, which is a root of a subtree
in the subset tree.
Output: A Boolean value indicating whether the subtree
rooted at S is pruned, along with an update of the best
subset found so far (stored in S_A).
// Terminal Step: S is a terminal node
if |S|= k then do
    if Dist(S) < min_dis then return true
    compute P_e(S)
    if P_e(S) < min_P_e then do
        min_P_e ← P_e(S)
        min_dis ← THRESH(min_P_e)
        S_A ← S        // update best subset so far
    end
    return false
end
// Recursive Step: S is an inner node
left ← (highest feature index in S)+1, and 1 if S = ∅
right ← n − k + |S| + 1
for i = left to right do
    if scan_tree(S ∪ {i}) = true
        then break        // quit loop
end
if i = left then return true // pruned subtree rooted at S
              else return false
```

Figure 5: Feature Selection Algorithm.

We claim that when *scan_tree* terminates, $S_A$ is one of the subsets $S^*$ having the minimal $P_e$. For this claim not to hold, all $S^*$ subsets would have had to be pruned at some iteration. This is not possible since $min\_dis$ was calculated according to a subset $S_A$ for which $P_e(S_A) \geq P_e(S^*)$, and therefore $\text{THRESH}(P_e(S^*)) \geq \text{THRESH}(P_e(S_A))$. Thus for every $S^*$, $Dist(S^*)$ must be greater than $min\_dis$, and therefore each $S^*$ is evaluated by the algorithm at some iteration. When the first $S^*$ subset is evaluated, its minimal $P_e$ is recorded, it is stored in $S_A$, and it stays in $S_A$ till termination.

This analysis relies on the fact that the subsets' $P_e$ can be computed exactly. This is possible in some cases, such as when the same covariance matrix is used for both classes, which enables exact computation of $P_e$ from the Mahalanobis distance (Cover & Van Campenhout, 1977). Generally, the only way to obtain the exact value of $P_e$ is by multidimensional integration; an insurmountable task even for a few dimensions. We therefore resort to estimating $P_e$.

### 6.1 Estimation of $P_e$

The estimation $\hat{P}_e(S)$ of $P_e(S)$ can be done using a set of $N$ data points, assumed to be selected i.i.d, for which the true class labels are known. If the class distribution parameters are known, the $N$ data points can be randomly generated according to the class prior distributions $p_{\omega_i}$ and the class conditional distributions $p(\vec{x}|\omega_i)$. This is known as a *Monte Carlo* method. Using the features in every subset $S$ examined, the classifier is tested on these $N$ points. The method for determining $N$ for each subset is explained below.

The estimation process of $\hat{P}_e(S)$ is viewed as a repeated series of $N$ *Bernoulli* experiments with an unknown parameter $\theta_e$, which is the error $P_e$ being estimated. The sum $Y$ of the $N$ experiments is distributed $Binomial(N, \theta_e)$, i.e., $P(Y=y)=\binom{N}{y}\theta_e^y(1-\theta_e)^{N-y}$. Assuming the $Beta(\alpha, \beta)$ distribution as a prior of $\theta_e$, the posterior of $\theta_e$ after viewing the outcome of $N$ experiments in which there were $a$ misses and $b$ correct classifications ($a+b=N$), is $Beta(\alpha+a, \beta+b)$ (Lee, 1997, p. 77). We model a lack of prior knowledge about $\theta_e$ with the hyper parameters $\alpha, \beta \to 0$.

After $N$ trials, with $a$ misses and $b$ correct classifications, the MAP-estimate $\hat{P}_e(S)$ equals $\frac{a+\alpha}{\alpha+\beta+N}$, which for small values of $\alpha, \beta$ approximately equals the ML-estimate $\frac{a}{N}$.

We now present a method for dynamically selecting $N$ for every subset evaluated. Consider two subsets, $S_1$ and $S_2$, the $P_e$ values of which we need to compare. We want our comparison to be $\varepsilon$-accurate with probability $1 - \delta$. That is, if $P_e(S_1) > P_e(S_2)$, we require that

$$Prob\{\hat{P}_e(S_1) \cdot (1+\varepsilon) > \hat{P}_e(S_2)\} > 1 - \delta \quad (6)$$

To ensure that Eq.(6) holds, it suffices that for $i=1,2$,

$$Prob\left\{\frac{\hat{P}_e(S_i)}{\sqrt{1+\varepsilon}} < P_e(S_i) < \hat{P}_e(S_i) \cdot \sqrt{1+\varepsilon}\right\} > 1 - \delta/2 \quad (7)$$

If Eq.(7) holds for both subsets, then the combined difference between the estimates for $P_e(S_1)$ and $P_e(S_2)$ and their true values is at most $1+\varepsilon$ with probability at least $1-\delta$. Assuming $\alpha, \beta \to 0$, if in the estimation process $N$ test points are used, with $a$ misses and $b$ correct classifications, then $\hat{P}_e(S_i) = \frac{a}{N}$, and given the $N$ test points, $\theta_e$ is distributed $Beta(a,b)$ a posteriori. Therefore, Eq.(7) holds when $a$ and $N$ satisfy:

$$\int_{\theta_e=\frac{a}{N}/\sqrt{1+\varepsilon}}^{\frac{a}{N}\cdot\sqrt{1+\varepsilon}} Beta(N \cdot \theta_e, N \cdot (1-\theta_e)) \, d\theta_e > 1 - \delta/2 \quad (8)$$

Note that the cdf values of the Beta distribution are easily available through standard math libraries. The estimation process draws $N$ i.i.d points, until Eq.(8) holds. For small $P_e$ values $N$ can be rather large. For instance, when $\varepsilon = 0.1$, $\delta = 0.1$, and $P_e = 0.01$, the required number of samples is $N = 168915$, whereas $N = 6833$ when $P_e = 0.2$, with the same $\varepsilon, \delta$. Small values of $\varepsilon$ and $\delta$ also increase $N$. For $P_e = 0.2$, $\varepsilon = 0.1$ and $\delta = 0.01$, the required number of samples is $N = 13974$. If $\varepsilon = 0.01$, $N$ rises to 1966080.

## 6.2 Comparing Branch and Bound Algorithms

We now compare our algorithm with another Branch and Bound algorithm for feature selection due to Narendra & Fukunaga (1977). NF's algorithm selects a globally optimal feature subset with respect to any monotonic criterion, such as Bhattacharyya distance, divergence, or $P_e$. To select a globally optimal feature subset of size $k$ from $n$ features, NF construct a *solution tree*, with the root being the full set of $n$ features. The successors of a node are created by removing a single feature from the node. This creates a tree with $n-k$ levels, the terminal nodes being all subsets of $\binom{n}{k}$ features. The solution tree is scanned in a depth-first fashion using the optimizing criterion value itself of the optimal subset encountered so far as the bound for pruning other nodes. For example, if the optimizing criterion is $P_e$, and the node $S$ examined has a $P_e$ larger than the current bound $b$ (the currently lowest $P_e$), the subtree rooted at $S$ is pruned. This pruning is justified whenever the optimizing criterion is monotonic which, in our example of $P_e$, means that for every subset $S'$ of $S$, if $P_e(S) > b$ so is $P_e(S')$.

Both NF's algorithm and ours perform a depth-first search on a tree to find the optimal subset. However, there are significant differences between these algorithms.

**Bounding and pruning methods** - When searching for the feature subset with the lowest $P_e$, our algorithm uses distance measures to prune the tree, whereas NF's algorithm uses $P_e$ itself for pruning. In our approach, in order to prune a branch we just compute its distance, a much simpler task.

**Robustness with large feature sets** - The estimation of $P_e$ done by our algorithm is for subsets of exactly $k$ features. NF's algorithm needs to compute $P_e$ for subsets ranging in size from $k$ to $n$, which carries a computational burden. In addition, using many features often yields very low $P_e$ values which requires larger sets of points for their estimation. This makes NF's algorithm less practical for high-dimensional datasets.

**Extent of pruning** - When a node is below the bound, both its subtree and its siblings' subtrees are pruned. In NF's algorithm only the node's subtree is pruned.

**Target goal** - NF's algorithm is designed to find subsets that optimize various monotonic criteria other than $P_e$. Our algorithm is aimed specifically at finding a subset with the lowest $P_e$, based on independent features. This specialization translates to improved performance.

## 7 Experimental Results

Section 7.1 describes synthetic data simulations designed to study the algorithm's performance. Section 7.2 offers a comparison of the performance of our algorithm and the one due to Narendra & Fukunaga (1977). Section 7.3 reports results on real data.

### 7.1 Synthetic Data

Our experiments in this section examine the influence of the features' distance distribution on the algorithm's performance. In the extreme case when all features have the same distance, no subset will have a distance below the threshold, and therefore no pruning will occur. On the other hand, when there are few features with high distances, and many with low distances, there is extensive pruning of the tree.

The experiments were conducted as follows. We repeatedly created datasets of $n$ features with different proportions of high distanced features (the distance of each feature being drawn uniformly form [0.15-0.3]), and low distanced features (drawn uniformly from [0-0.15]). Once a distance was chosen for feature $i$, the class parameters ($\mu_{1i}, \sigma_{1i}, \mu_{2i}, \sigma_{2i}$) were set to have that distance in the following manner: The parameters of the first class were always $\mu_{1i}=0$ and $\sigma_{1i}=1$. In Eq.(3) for the Bhattacharyya distance, there are two factors which contribute to the distance: $\frac{1}{4}\frac{(\mu_{2i}-\mu_{1i})^2}{\sigma_{1i}^2+\sigma_{2i}^2}$, which depends both on the class means and variances, and $\frac{1}{2}\ln\frac{\sigma_{1i}^2+\sigma_{2i}^2}{2\sigma_{1i}\cdot\sigma_{2i}}$, which depends only on the class variances. A weight was drawn uniformly from [0,1] to indicate to what extent the first factor contributes to the feature distance, enabling us to set $\mu_{2i}$ and $\sigma_{2i}$ to unique values. The class prior probabilities in all the experiments were $p_{\omega_1}=p_{\omega_2}=0.5$.

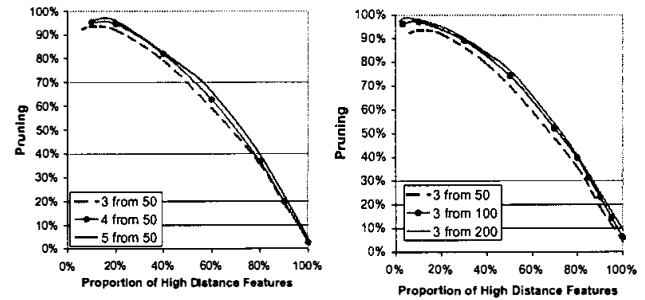

Figure 6: Pruning with synthetic data.

Figure 6 shows our algorithm's output in terms of subset pruning, i.e. the proportion of the $\binom{n}{k}$ subsets for which no evaluation was necessary because they had a distance which was less than the threshold. The algorithm was run with the parameters $\varepsilon=0.1$, $\delta=0.1$, and different values of $n$ (total number of features) and $k$ (subset size). Different proportions of high distanced features were also used. The best pruning is achieved when there are relatively few features with high distances. As the proportion of the features with high distances rises, the pruning rate decreases.

Figure 7 depicts results of additional experiments we ran to determine the tradeoff between the accuracy of the $P_e$ estimations and the quality of the subset chosen by the algorithm. A set of 20 different datasets was generated as described above, with parameters $n=50$, $k=3$, and half of the features having a high distance. The algorithm was run on the 20 datasets using different values of $\varepsilon$ and $\delta$.

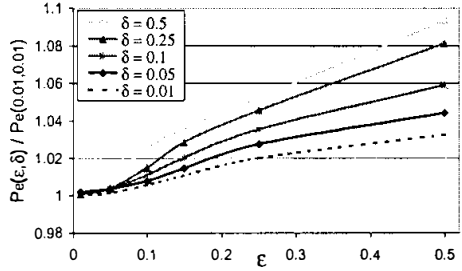

Figure 7: The average $P_e$ of the subsets selected by the algorithm when different values of $\varepsilon$ and $\delta$ were used, compared to $P_e$ achieved with parameters $\varepsilon = 0.01$ and $\delta = 0.01$, which gives the maximal accuracy.

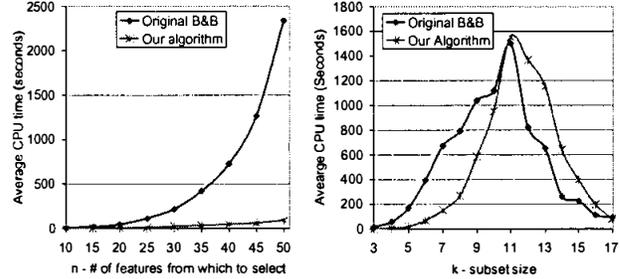

Figure 8: The average CPU time required for both algorithm to select feature subsets. The graphs are for a selection of 3 features from $n$ (*left*), and a selection of $k$ features from 20 (*right*). The CPU times were measured on a desktop PC with an Intel P-IV 2.4GHz processor.

## 7.2 Comparison to Other Algorithm

We implemented the Branch and Bound feature selection algorithm of Narendra & Fukunaga (1977). We used the same methods for estimating the Bayesian classification error as described in Section 6.1, including the early termination of the error estimation when possible.

NF's algorithm spends much of its efforts estimating $P_e$ values for inner nodes of its solution tree. Our algorithm only estimates the errors of the terminal nodes in the tree. Therefore, looking at the pruning rate of the terminal nodes is not a good criterion by which the algorithms should be compared. Instead, we chose to compare the CPU time it took each algorithm to return its choice of a feature subset.

Both algorithms were run on the same synthetic datasets, with each feature having a Bhattacharyya distance drawn uniformly [0,0.2], and its parameters set in the manner described in Section 7.1. Both algorithms were run with the parameters $\varepsilon = 0.1$, $\delta = 0.1$. Figure 8 depicts the CPU time required by both algorithms to return an optimal subset of $k$ features from $n$, for various values of $n$ and $k$. Note that both algorithms usually returned the same subset of features, or very similar subsets, with very close $P_e$ values.

It is apparent from Figure 8 that our algorithm performs much faster when $k$ is much smaller than $n$. The main reason for this phenomenon is that the only error estimations our algorithm performs are for terminal nodes of $k$ features in the subset tree. NF's algorithm, on the other hand, evaluates many inner nodes in the tree, nodes which represent subsets of between $n$ to $k$ features. Before NF's algorithm reaches a terminal node, it must estimate errors for all $n-k$ subsets on the path from the root to the terminal node. This carries a heavy computational burden. In addition, larger feature subsets usually have low $P_e$'s, so they require larger test samples for the error estimation.

When $k$ is larger, for instance $k > \frac{n}{2}$, NF's algorithm is faster than ours (see Figure 8, right). The main reason this happens is that for large values of $k$ NF's tree is shorter, with less intermediate nodes. Therefore, fewer $P_e$ computations are done, compared to the number required in NF's tree when $k$ is small.

## 7.3 Application to Gene Expression Data

We used our algorithm for feature selection in gene expression datasets (Slonim *et al.*, 2000). A typical gene expression dataset consists of several thousands of genes, with only a small subset of them being significant to distinction between classes. Using all the genes does not give good classification results, so feature selection is crucial in this domain. Along with the data's high dimensionality, comes a limited sample space. Usually gene expression datasets consist of less than 100 samples. The limited number of samples prohibits us from using our algorithm directly for feature selection. Instead, we use our algorithm as a heuristic, as explained below.

We examine two datasets, The AML/ALL distinction in the leukemia dataset of Golub *et al.* (1999) (7129 genes, 72 samples: 38 in the training set and 34 in the test set), and the ER+/ER- distinction in the breast cancer data of Gruvberger *et al.* (2001) (3389 genes, 58 samples: 47 in the training set and 11 in the test set). For each dataset we estimate the parameters $p(\vec{x}|\omega_i)$ and $p(\omega_i)$ from their observed probabilities in the training set. We then select the 300 genes with the highest Bhattacharyya distance, and use them as the input to our algorithm, which is run with parameters $\varepsilon, \delta = 0.1$.

During the algorithm's execution, every subset that is not pruned according to the minimal distance bound and its $P_e$ (which is calculated using the Monte Carlo method), is subjected to $N$ 5-fold cross validation tests (where $N$ is determined according to the $\varepsilon, \delta$ accuracy, similar to the method described in Section 6.1). Since $P_e$ is only used to determine the distance bound, its computation can be omitted for many subsets. This might yield slightly looser bounds, however it can ultimately reduce the overall running time (this is especially true when $P_e$ values are low and their computation requires many random points).

The best subset is chosen by the algorithm as the one with the lowest average cross validation error. We then test its ability to classify the samples in the test set. The results of these experiments are listed in Table 2.

Table 2: Classification and Pruning Results.

| DATASET | K | BEST SUBSET[1] | SUCCESS | PRUNING[2] |
|---|---|---|---|---|
| leukemia | 3 | 3, 4, 60 | 33/34 | 99.77% |
|  | 4 | 3, 4, 8, 60 | 33/34 | 99.91% |
|  | 5 | 2, 3, 4, 9, 60 | 34/34 | 99.57% |
| breast cancer | 3 | 1, 14, 88 | 8/11 | 99.25% |
|  | 4 | 1, 18, 44, 67 | 11/11 | 99.73% |
|  | 5 | 1, 8, 18, 44, 67 | 11/11 | 99.76% |

[1] Features are indexed according to their Bhattacharyya distance, feature 1 having the highest distance.

[2] The pruning is the percentage of the $\binom{300}{k}$ subsets that were not examined by the algorithm.

The classification success rates achieved by our method are comparable or better than the previously published classification rates for these datasets. This means that even though we examine a small fraction of the $\binom{300}{k}$ feature subsets, our method selects qualitative candidate subsets. Our method's high pruning rates make up for the overhead involved in the Monte Carlo $P_e$ calculations. For instance, without omitting any $P_e$ calculations, it takes our method 30 minutes to select the best subset of 3 from 300 features in the leukemia data, whereas an exhaustive search is 15 times slower, and yields an inferior subset which classifies only 28/34. Similar results were obtained with the breast cancer data, the subset chosen by an exhaustive search classifies 7/11. A possible explanation for this is that when all the subsets are examined in an exhaustive search, there is an increased risk of encountering a subset that is over-fitted to the training set.

The subsets selected by our algorithm contain features with relatively high Bhattacharyya distances (this is especially true in the leukemia data). This shows that the Bhattacharyya distance is a good measure for predicting a feature's contribution to successful classification.

We also examined greedy feature selection methods, such as selecting the features with the highest individual Bhattacharyya distances, or with the lowest individual $P_e$ values. The greedy methods give mixed results. They are successful with the breast cancer data. Like our method, they achieve perfect classification of its test set. However, with the leukemia data, the greedy methods do not succeed in selecting good feature subsets.

## 8 Conclusion

We presented a new Branch and Bound algorithm for feature selection aimed at finding a feature subset of a given size with the lowest Bayesian classification error. We use the Bhattacharyya distance, which is monotonic and additive (due to the independence assumption of our statistical model), to exclude many candidate feature subsets from the search process. Our algorithm performs a complete search, it examines all the subsets that can possibly have the lowest classification error. In many cases due to extensive pruning, the algorithm's search is far from being exhaustive. We described the Monte Carlo method we use for estimating the Bayesian classification error when its exact computation is not possible. A comparison of our implementation and the Branch and Bound algorithm due to Narendra & Fukunaga (1977) identifies the conditions under which our algorithm has superior performance. We also demonstrated how our method can be used successfully for feature selection in gene expression datasets.